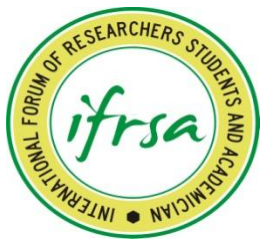

# IJGIP

# Hybrid Image Segmentation using Discerner Cluster in FCM and Histogram Thresholding

Firas A. Jassim
Management Information System Department, Irbid National University, 2600 Irbid - Jordan

**ABSTRACT**

Image thresholding has played an important role in image segmentation. This paper presents a hybrid approach for image segmentation based on the thresholding by fuzzy c-means (THFCM) algorithm for image segmentation. The goal of the proposed approach is to find a discerner cluster able to find an automatic threshold. The algorithm is formulated by applying the standard FCM clustering algorithm to the frequencies (y-values) on the smoothed histogram. Hence, the frequencies of an image can be used instead of the conventional whole data of image. The cluster that has the highest peak which represents the maximum frequency in the image histogram will play as an excellent role in determining a discerner cluster to the grey level image. Then, the pixels belong to the discerner cluster represent an object in the gray level histogram while the other clusters represent a background. Experimental results with standard test images have been obtained through the proposed approach (THFCM).

**Keywords**: Image thresholding; automatic threshold, fuzzy c-means; image histogram, THFCM

## 1. INTRODUCTION

Segmentation subdivides an image into its constituent regions or objects. The level to which the subdivision is carried depends on the problem being solved. That is, segmentation should stop when the objects of interest in an application have been isolated. Segmentation of nontrivial images is one of the most difficult tasks in image processing [6].

The most basic and important technique in image processing is image segmentation and it is necessary in pattern recognition. The basic idea behind segmentation is to segment an image into several clusters. According to the segmentation results, it is possible to identify regions of interest and objects in the original image. Recent work includes a variety of techniques like Fuzzy c-means (FCM) clustering algorithm which is a popular model widely used in image segmentation due to its good performance [3][4]. Several algorithms have been proposed to improve FCM in incorporating the spatial information of the image. Though many approaches existed, people are still very interested in developing the algorithms that can quickly and correctly segment an image. According to reference [5], the image segmentation approaches can be divided into four categories: thresholding, clustering, edge detection and region extraction. In this paper, a hybrid approach based on clustering and thresholding methods for image segmentation will be considered.

## 2. FUZZY C-MEANS

The fuzzy c-means (FCM) algorithm assigns pixels to each cluster by using fuzzy memberships. Let $X=(x_1, x_2,..,x_N)$ denotes an image with N pixels to be partitioned into c clusters, where $x_i$ represents multispectral (features) data. The algorithm is an iterative optimization that minimizes the cost function defined as follows [7]:

$$J = \sum_{j=1}^{N}\sum_{i=1}^{c} u_{ij}^{m} \left\| x_j - v_i \right\|^2$$

where $u_{ij}$ represents the membership of pixel $x_j$ in the $i$th cluster, $v_i$ is the $i$th cluster center, $\|.\|$ is a norm metric, and $m$ is a constant. The parameter $m$ controls the fuzziness of the resulting partition, and $m=2$ is used in this study. The cost function is minimized when pixels close to the centroid of their clusters are assigned high membership values, and low membership values are assigned to pixels with data far from the centroid. The membership function represents the probability that a pixel belongs to a specific cluster. In the FCM algorithm, the probability is dependent solely on the distance between the pixel and each individual cluster center in the feature domain. The membership functions and cluster centers are updated by the following:

$$u_{ij} = \frac{1}{\sum_{k=1}^{c}\left(\frac{\|x_j - v_i\|}{\|x_j - v_k\|}\right)^{\frac{2}{m-1}}}$$

and

$$v_i = \frac{\sum_{j=1}^{N} u_{ij}^m x_j}{\sum_{j=1}^{N} u_{ij}^m}$$

Starting with an initial guess for each cluster center, the FCM converges to a solution for $v_i$ representing the local minimum or a saddle point of the cost function [3]. Convergence can be detected by comparing the changes in the membership function or the cluster center at two successive iteration steps.

### 3. HISTOGRAM THRESHOLDING

Threshold is one of the widely methods used for image segmentation. It is useful in discriminating foreground from the background. By selecting an adequate threshold value $T$, the gray level image can be converted to binary image. The binary image should contain all of the essential information about the objects of interest. The advantage of obtaining first a binary image is that it reduces the complexity of the data and simplifies the process of recognition and classification. The most common way to convert a gray-level image to a binary image is to select a single threshold value $T$. Then all the gray level values below this $T$ will be classified as black (0), and those above $T$ will be white (1). The segmentation problem becomes one of selecting the proper value for the threshold $T$. A frequent method used to select $T$ is by analyzing the histograms of the type of images that want to be segmented. The ideal case is when the histogram presents only two dominant modes. In this case the value of $T$ is selected as the valley point between the two modes. In real applications histograms are more complex, with many peaks and not clear valleys, and it is not always easy to select the value of $T$. Automatically selected threshold value for each image by the system without human intervention is called an automatic threshold scheme [10]. This is requirement the knowledge about the intensity characteristics of the objects, sizes of the objects, fractions of the image occupied by the objects and the number of different types of objects appearing in the image [1].

The key parameter in the thresholding process is the choice of the threshold value. Several different methods for choosing a threshold exist; users can manually choose a threshold value, or a thresholding algorithm can compute a value automatically, which is known as automatic thresholding. A simple method would be to choose the mean or median value, the rationale being that if the object pixels are brighter than the background, they should also be brighter than the average. In a noiseless image with uniform background and object values, the mean or median will work well as the threshold, however, this will generally not be the case. A more sophisticated approach might be to create a histogram of the image pixel intensities and use the valley point as the threshold. The histogram approach assumes that there is some average value for the background and object pixels, but that the actual pixel values have some variation around these average values [10]. However, this may be computationally expensive, and image histograms may not have clearly defined valley points, often making the selection of an accurate threshold difficult. One method that is relatively simple, does not require much specific knowledge of the image, and is robust against image noise, is the following iterative method.

### 4. HYBRID THRESHOLDING AND FCM (THFCM)

The general principle of the techniques presented in this paper is to find an automatic threshold that can be used to separate objects from the background. The traditional concept in thresholding is to find vertical line which cuts the x-axis. The left area of the line represents either an object or background while the right area is the converse. There are recent works to introduce multi-level thresholding [9].

Firstly, a histogram for the image can be established which contains frequencies from 0-255 for all image pixels. After that, smoothing the histogram which will introduce a smooth curve mimics the bars of the histogram. Mathematically speaking, every pixel on the curve representing an ordered pair in the cartesian coordinates (x,y) where the x-value lies between [0-255] while the y-value is for the frequencies for each pixel. The main important principle for the researchers in thresholding is to find a threshold that cuts the x-axis by a vertical line while, in this paper, a novel techniques can be used instead. Secondly, a fuzzy c-means algorithm can be applied on the smoothed curve (y-values) instead on the traditional fuzzy c-means that had been applied to the x-values. Many researchers have discussed the problem of the number of clusters [2][8], but according to [8] the suitable number of clusters in the proposed method is three clusters. The cluster which has the highest frequency (maximum) in the y-value is the discerner cluster that may be treated as a threshold to separate all the frequencies (y-values) inside that cluster into an object (1). While the other two clusters, from the whole three clusters, may be treated as background (0) such as:

$$g(y) = \begin{cases} 1 & , y \in c_i \\ 0 & , y \in c_j \end{cases}, i \neq j, j \in [1..3]$$

One of the important characteristics of an image is that neighboring pixels are highly correlated. In other words, these neighboring pixels possess similar feature values, and the probability that they belong to the same cluster is great.

The proposed method in this paper aims to overcome some of the limitations of the existing method, i.e. determining a threshold. In fact, the discerner cluster is defined automatically and it seems to be fair to accommodate the highest peak values in the smoothed histogram curve. Consequently, the determination of an automatic threshold may be computed without any previous known or preparation for the threshold.

## 5. EXPERIMENTAL RESULTS

In this section, the results of the thresholding by fuzzy c-means (THFCM) algorithm are presented through figures 1, 2 and 3. The application of the proposed approach results with standard test images has been obtained. It can be seen that three standard test images have been applied, (F16, Lena and Baboon). Graphically speaking, each image has a smooth histogram with circles representing the discerner cluster which may be treated as an automatic threshold for image segmentation.

For each histogram from the previous histograms, it can be seen that the circles on the smoothed histogram were found on almost high peaks. These circles represent the discerner cluster obtained by FCM from three clusters. The methodology for choosing the discerner cluster depends on the highest y-value (peak) on the histogram.

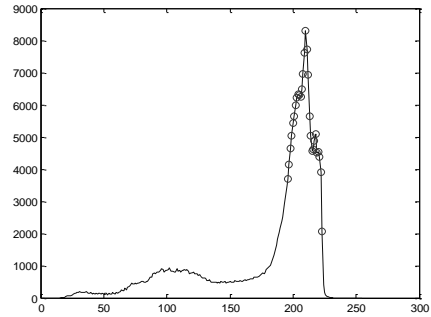
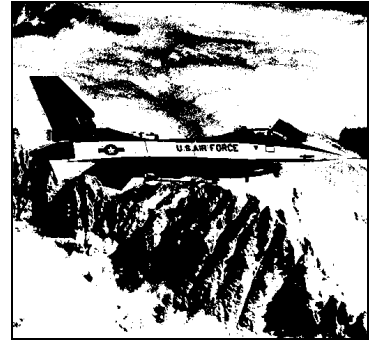

*Figure (1) Discerner cluster for F16 image*

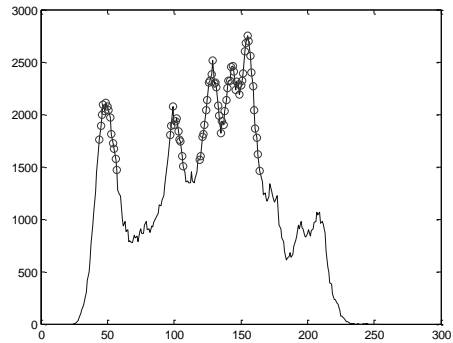
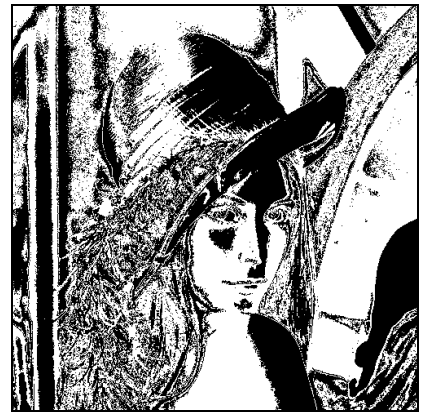

*Figure (2) Discerner cluster for Lena image*

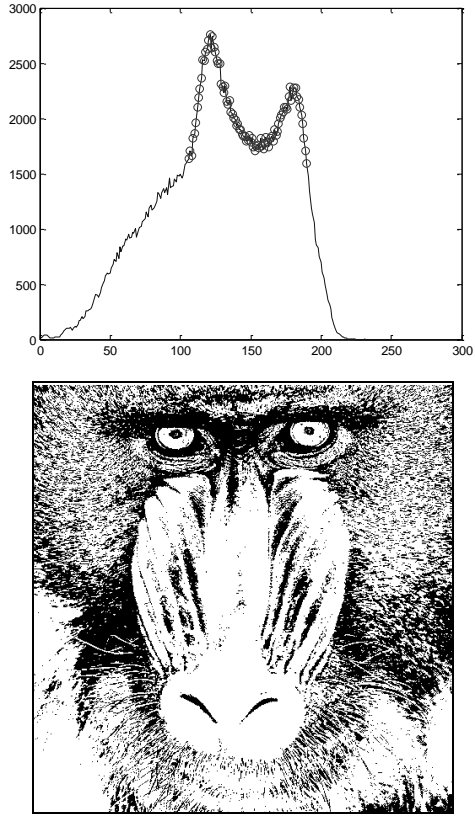

*Figure (3) Discerner cluster for Baboon image*

## 6. CONCLUSION & FUTURE WORK

In this paper, an automatic histogram threshold approach based on the discerner cluster by fuzzy c-means has been obtained. A novel measure of fuzziness is presented. This work overcomes some limitations of the traditional thresholding method concerning the definition of the initial threshold cut point.

Firstly, one of the main recommendations for the future work is that variable cluster-size to determine the discerner cluster instead of fixing the number of clusters in FCM into three.

Secondly, THFCM may be applied to color images instead of the grey-level images that have been used in the research.

Thirdly, a hybridizing between thresholding and k-means may be used instead of the hybridizing used in this work which was between FCM and Thresholding.

The segmentation results are quit acceptable for image segmentation but these results are not more accurate than recent work but it serves as novel approach in image segmentation using hybridizing between FCM and thresholding that can be used as a scheme embedded into other methods to focus on the discerner cluster instead of the whole FCM clusters.